\documentclass[9pt,conference]{IEEEtran}
\usepackage{graphicx}     % essential
\usepackage{stfloats}     % lets a figure* go at bottom if you need [!b]
\usepackage{booktabs}   % \toprule, \midrule, \bottomrule, \cmidrule
\usepackage{multirow}   % \multirow for the diagonal cell spanning two header rows
\usepackage{tabularx}   % auto‑stretch columns to full width
\usepackage{siunitx}    % optional: align decimals, e.g. S[table-format=2.2]
\usepackage{booktabs,tabularx,makecell}   % in the preamble
\usepackage{array,booktabs,tabularx}  % add 'array' for the >{} spec
\usepackage{booktabs,tabularx,makecell,array}   % <-- add 'makecell' +
\usepackage{authblk} % for \email command
\usepackage{booktabs}   % For horizontal lines
\usepackage{amssymb}    % For arrow symbols'array'
\usepackage{placeins}   % For preventing floats from passing the barrier

\usepackage[preprint]{waspaa25}

\usepackage{bm} % for bold math symbols (incl. Greek letters) with \bm{}

\title{Unveiling the Best Practices for Applying Speech Foundation Models to Speech Intelligibility Prediction for Hearing-Impaired People}

\name{Haoshuai Zhou$^{1}$,
      Boxuan Cao$^{1}$,
      Changgeng Mo$^{1}$,
      Linkai Li$^{*,1,2}$
      Shan Xiang Wang$^{*,2,3}$}
\address{$^{1}$Orka Labs Inc., China \;
$^{2}$Electrical Engineering, Stanford University, United States\\
$^{3}$Materials Science and Engineering, Stanford University, United States \\
\texttt{\{haoshuai,dicky.mo,boxuan.cao\}@hiorka.com, \{linkaili,sxwang\}@stanford.edu}
\thanks{$^{*}$Corresponding authors}
}

\begin{document}

\maketitle

\begin{abstract}
Speech foundation models (SFMs) have demonstrated strong performance across a variety of downstream tasks, including speech intelligibility prediction for hearing-impaired people (SIP-HI). However, optimizing SFMs for SIP-HI has been insufficiently explored. In this paper, we conduct a comprehensive study to identify key design factors affecting SIP-HI performance with 5 SFMs, focusing on encoder layer selection, prediction head architecture, and ensemble configurations. Our findings show that, contrary to traditional use-all-layers methods, selecting a single encoder layer yields better results. Additionally, temporal modeling is crucial for effective prediction heads. We also demonstrate that ensembling multiple SFMs improves performance, with stronger individual models providing greater benefit. Finally, we explore the relationship between key SFM attributes and their impact on SIP-HI performance. Our study offers practical insights into effectively adapting SFMs for speech intelligibility prediction for hearing-impaired populations.
\end{abstract}

\section{Introduction}
\label{sec:intro}
Speech intelligibility (SI) is a key performance indicator for hearing aid algorithms. To automate SI prediction for hearing‑impaired listeners (SIP‑HI), researchers first run subjective listening tests and then derive objective metrics from the collected scores \cite{Barker2022The1C, Barker2024The2C, Kates2014TheHS}. However, existing SIP‑HI datasets are relatively small, often with fewer than 30 listeners\cite{Barker2022The1C,Barker2024The2C,Plyler2021EffectOH,Monaghan2017AuditoryIM,Miles2022MeasuringSI}. Thus, training deep neural models from scratch on them typically overfits and delivers sub‑optimal accuracy. 

Recently, research has shifted toward developing general purpose models, which are called foundation models, that are trained on broad and large-scale data and finetuned for downstream tasks \cite{Bommasani2021OnTO}. This transfer‑learning based paradigm has demonstrated its effectiveness in boosting generalization in data‑limited settings \cite{Baevski2020wav2vec2A, Hsu2021HuBERTSS,Brown2020LanguageMA}. In speech area, adapting pretrained speech foundation models, such as wav2vec2.0 \cite{Baevski2020wav2vec2A} and HuBERT \cite{Hsu2021HuBERTSS} has yielded state‑of‑the‑art performance on downstream tasks, including keyword spotting, speaker identification and intent classification\cite{Yang2021SUPERBSP}. 

Consequently, researchers have begun adapting speech foundation models (SFMs) for SIP‑HI and have achieved encouraging results. In the 2nd Clarity Prediction Challenge (CPC2)\cite{Barker2024The2C}, the two top‑ranked systems combine a frozen SFM backbone with an auxiliary prediction head, where both heads ingest representations from every encoder or decoder layer of the SFM and fuse them through learned weights or attention mechanism\cite{Cuervo2024SpeechFM,Mogridge2024NonIntrusiveSI}. Nevertheless, evidence from other speech tasks, such as {speech quality assessment, intent classification and emotion recognition shows that not all SFM layers contribute equally to downstream performance\cite{Waheed2024WhatDS,Tamm2023AnalysisOX}, which cautions against indiscriminate layer aggregation and calls into question whether the prevailing “use‑all‑layers” strategy is truly optimal for SIP‑HI. Besides, previous research offer only limited ablation of prediction‑head design, leaving it unclear how the prediction head structure actually drive performance gains, which in turn makes systematic refinement of SIP‑HI architectures more challenging. Furthermore, existing methods lack analysis on the criterion of selecting SFMs for ensembling to maximize performance.

To address these questions, this paper aims to provide a recipe for effectively leveraging SFMs for SIP-HI. Our investigation is built around an adapter-based framework, where the SFM is used as a frozen backbone and a lightweight prediction head is fine-tuned for the SIP-HI task. We experiment with five state-of-the-art SFMs and perform a comprehensive layer-wise analysis to explore how encoder layer depth influences SIP-HI performance. In addition, we evaluate multiple prediction heads of varying complexity to understand how adapter architectural choices affect model accuracy. Moreover, we conduct experiments on ensembling different SFMs to investigate which combinations yield higher performance and how each SFM contributes to the ensemble. Our results demonstrate that:

1. Encoder layer selection is crucial for SIP-HI performance. In nearly all SFM and prediction head configurations, the best-performing models rely on features from a single encoder layer, rather than self-learned weighted combinations of all layers. Moreover, there is no consistent pattern across SFMs regarding which encoder layer yields the best results, emphasizing the need for layer-wise evaluation when adapting a new SFM to SIP-HI.

2. Prediction head design is important, with temporal modeling capability proving more influential than layer fusion capability or embedding dimensions.

3. Ensembling different SFMs can significantly improve performance compared to individual models, with stronger SFMs being more effective candidates for the ensemble.

\section{Experimental Setup}
\subsection{Dataset}
We run all experiments on the CPC dataset \cite{Barker2022The1C,Barker2024The2C}. The dataset contains 18 different hearing aid processing systems and 27 hearing-impaired listeners, and in total has 13,126 speech clips under diverse acoustic scenes together with intelligibility scores collected from listening experiments. We repartition the official development set into disjoint training, validation, and test set and ensure there is no listeners overlap. We also adopt the three-fold practice as in \cite{Barker2024The2C} to reduce score variance.

\subsection{Speech Foundation Models}
In our experiments, we evaluate 5 state-of-the-art speech foundation models that cover a range of architectures and training regimes:

\textbf{Canary} \cite{puvvada2024less} is a SFM trained on about 86K hour of multilingual automatic speech recognition (ASR) and automatic speech translation (AST) data. Its architecture couples a FastConformer \cite{Rekesh2023FastCW} with a Transformer \cite{Vaswani2017AttentionIA} decoder. We use the Canary-1b version checkpoint.

\textbf{Parakeet} is a SFM which is architecturally similar to Canary but is trained solely for English ASR on 64K hour speech data. In our experiment, we use the Parakeet-tdt-1.1b version which brings about improved inference efficiency and robustness to noisy speech \cite{Xu2023EfficientST}.

\textbf{Whisper}\cite{Radford2022RobustSR} is a vanilla Transformer-based encoder–decoder foundation model. It is pre-trained on a range of speech-related tasks—including ASR, AST, language identification (LID) and voice activity detection (VAD)—using an initial dataset of 68K hours of weakly labeled web audio, later expanded to over 5M hours through pseudo-labeling. Whisper demonstrates strong task robustness, along with excellent zero-shot and few-shot generalization, particularly in multilingual and low-resource settings. In this paper, we use Whipser\_large-v3.

\textbf{OWSM}\cite{Peng2023ReproducingWT} is a open-source foundation model developed to reproduce Whisper using only publicly available data and toolkits. The initial version replicates Whisper’s original architecture, while later versions explore alternative encoder designs. In our experiments, we use OWSM\_v3.1\_ebf\cite{Peng2024OWSMVB}, which replaces the standard Transformer encoder with an E-Branchformer \cite{Kim2022EBranchformerBW}, offering improved accuracy and faster inference without additional training data.

\textbf{Phi-4-Multimodal}\cite{Abouelenin2025Phi4MiniTR} is a large-scale multimodal foundation model designed to handle text, vision, and speech/audio inputs within a unified architecture. In our experiments, we use the phi-4-multimodal-instruct version and focus exclusively on the speech modality. We provide the model with audio-only inputs and use an instruction-style prompt to perform transcription. The exact prompt used is:
\texttt{\textless|user|\textgreater\textless|audio\_1|\textgreater Transcribe the audio clip into text.\textless|end|\textgreater\textless|assitant|\textgreater}. This prompt leverages the model’s instruction-following capability and allows us to extract its internal speech representations for downstream tasks.

In the remainder of this paper, for brevity, we use Canary, Parakeet, Whisper, OWSM and Phi-4 to refer to the specific version SFMs employed in our experiments.

\begin{figure}
    \centering
    \includegraphics[width=1.0\linewidth]{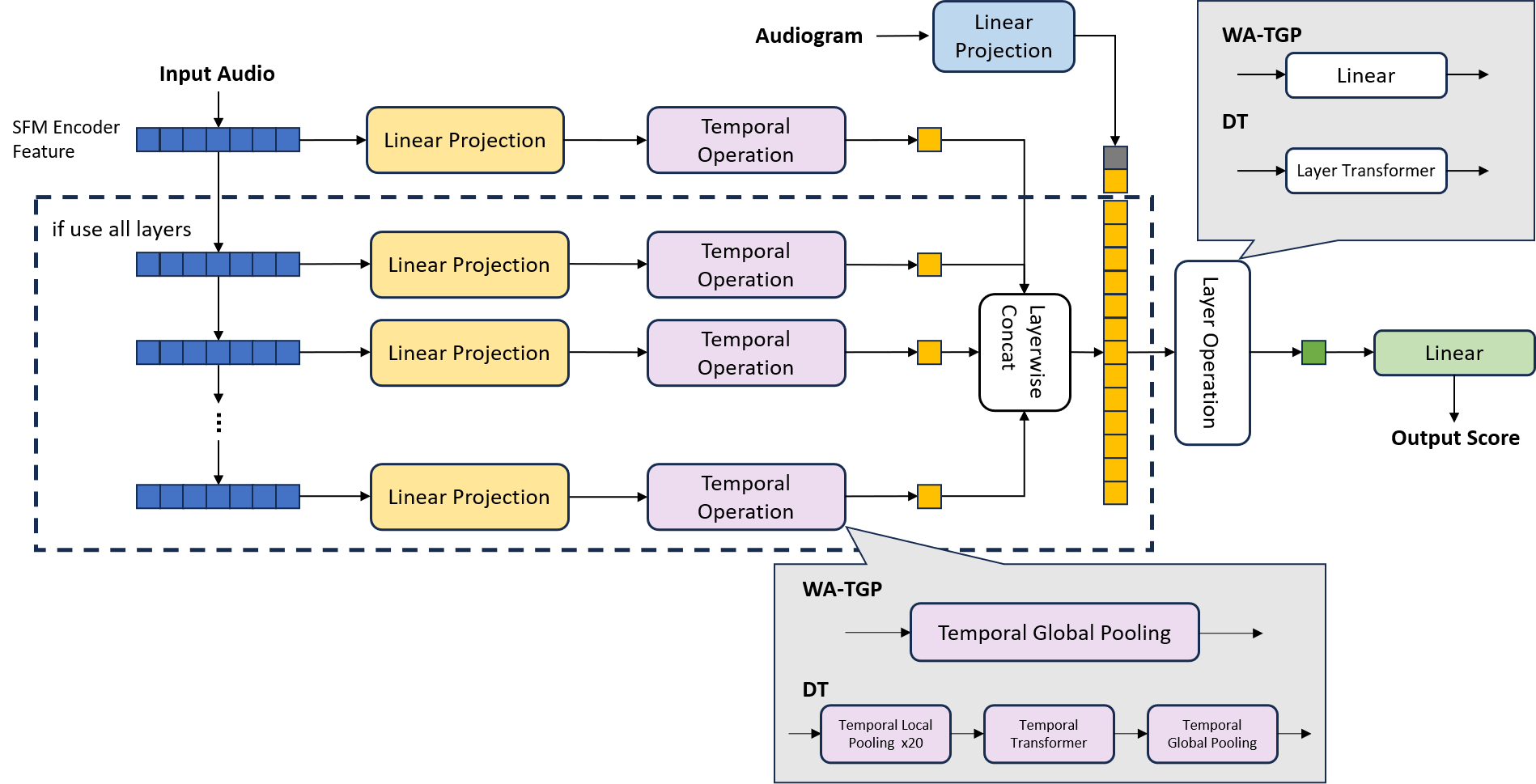}
    \caption{Two prediction head structures used: WA-TGP and DT. Channel dimensions are omitted for illustration. The same color indicates shared weights.}
    \label{fig:model_structure}
\end{figure}

\subsection{Encoder Feature Layer Selection}
One limitation we observe in prior SIP-HI methods is the default strategy of using all encoder or decoder layer outputs from the SFM as input to the intelligibility predictor\cite{Cuervo2024SpeechFM,Mogridge2024NonIntrusiveSI}. This implicit reliance on a learned layer-weighting or attention mechanism raises the question of whether certain layers are inherently more informative for the SIP-HI task and complementary information can actually be learned given all layers. To investigate this, we evaluate and compare two prediction head architectures: WA-TGP (Weighted-Average with Temporal Global Pooling) and DT(Double Transformer), as illustrated in \cref{fig:model_structure}.

The first is a simple weighted average-based structure adpopted from \cite{Gong2023WhisperATNA}. In this architecture, the encoder output from each SFM layer first undergoes a channel-wise linear projection and temporally global average pooling. Then a set of learned scalar weights is applied to combine layer-wise features. When using one encoder layer, the combination includes the encoder layer feature and a linearly projected audiogram feature. When using multiple layers, all encoder features are combined with the audiogram feature. The resulting weighted feature is averaged across the left and right ears, followed by another linear layer to predict the intelligibility score. While this learned weighting mechanism allows the model to emphasize relevant layers, the architecture is relatively shallow, and the initial brute-force global temporal pooling risks discarding useful temporal patterns early in the process.

To address this, we also add a transformer-based architecture, which is utilized in the winning submission of the CPC2\cite{Cuervo2024SpeechFM}. After linear projection, for temporal operations, encoder features from each layer or all layers are first pooled by a factor of 20, then passed through a temporal transformer to preserve sequential context and then finally temporally global averaged. A layer-wise transformer is then used to dynamically combine information across layers and the audiogram-based features. As in the previous setup, the left and right channel features are averaged, and a linear layer is used to produce the final intelligibility score.

For the above methods, we set the dimensions of the projected feature, as well as the internal and output dimensions of the transformer, all to 384. 

\subsection{Prediction Head Comparison}
In addition to exploring encoder layer selection, we argue that a deeper analysis of the prediction head architecture is also essential for understanding and improving SIP-HI performance. Building on the CPC2 champion design\cite{Cuervo2024SpeechFM}, we investigate two key factors: (i) how temporal information is extracted, and (ii) how layer-level features are fused.

To study these aspects, we design three different prediction heads. The first two are the previously described WA-TGP and DT structures. These represent two extremes in complexity: WA-TGP offers a simple approach to both temporal and layer fusion, while DT uses temporal and layer-wise transformers for more expressive modeling.

To bridge the gap between these two extremes, we introduce a third architecture: Weighted-Average with Temporal Transformer (WA-TT). After channel-wise projection, this structure first applies a temporal transformer, as in the DT model, to extract rich sequential information. It then fuses the layer-wise features using a weighted average, following the same strategy as WA-TGP. WA-TT thus represents a moderately complex design that isolates the impact of temporal modeling while simplifying layer aggregation.

For all three prediction head designs, we sweep across four embedding dimensions: 192, 384, 768, and 1536. For WA-TGP, this corresponds to the linear projection dimension. For temporal transformer-based models, the same values are used for the linear projection layer as well as the internal and output dimensions of the transformer blocks. In the case of WA-TGP and DT, we use the best-performing encoder layer from the layer selection experiment. For WA-TT, we hypothesize that its optimal layer choice should be similar to that of WA-TGP. Therefore, for simplicity, we use the same layer configuration for WA-TT as for WA-TGP.

\begin{figure*}[!t]  % figure* spans both columns
  \centering
  % First row of 5 images
  \includegraphics[width=.19\textwidth]{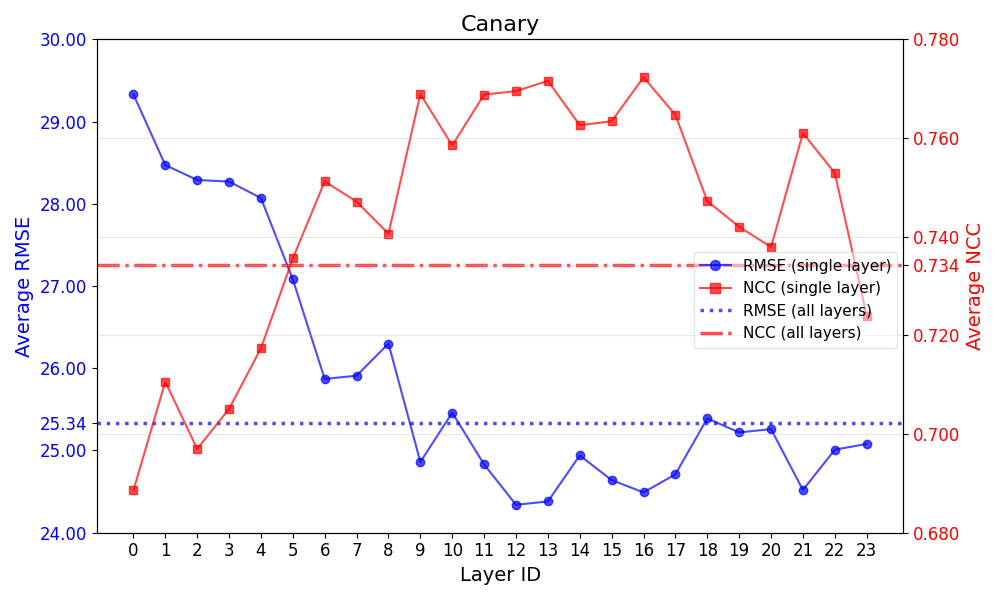}\hfill
  \includegraphics[width=.19\textwidth]{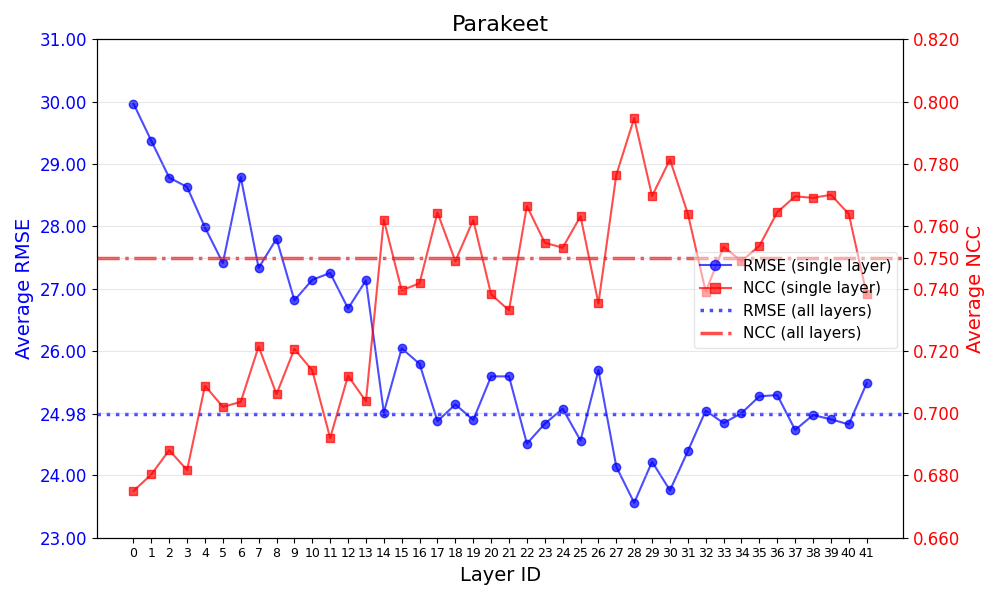}\hfill
  \includegraphics[width=.19\textwidth]{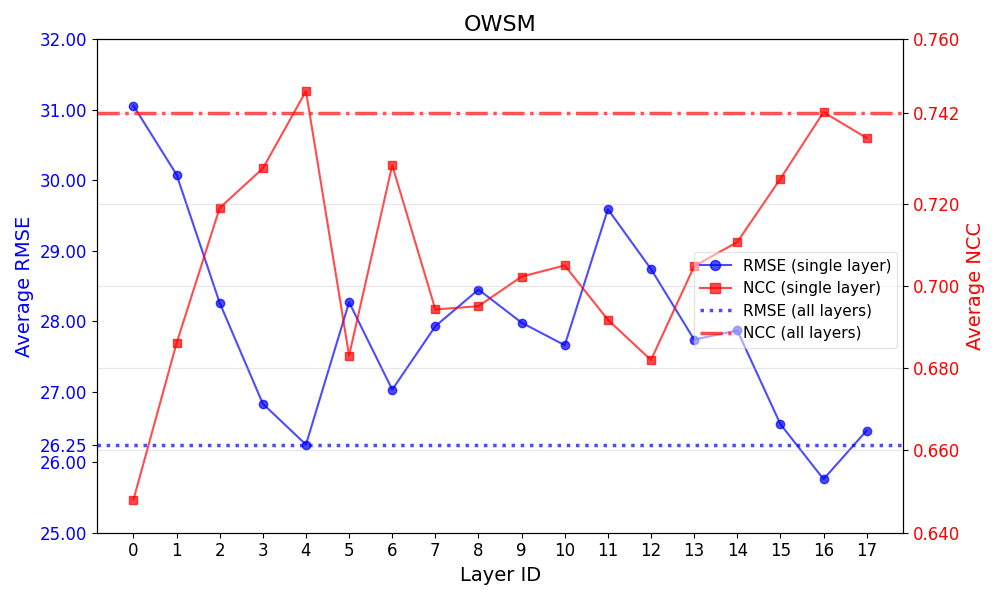}\hfill
  \includegraphics[width=.19\textwidth]{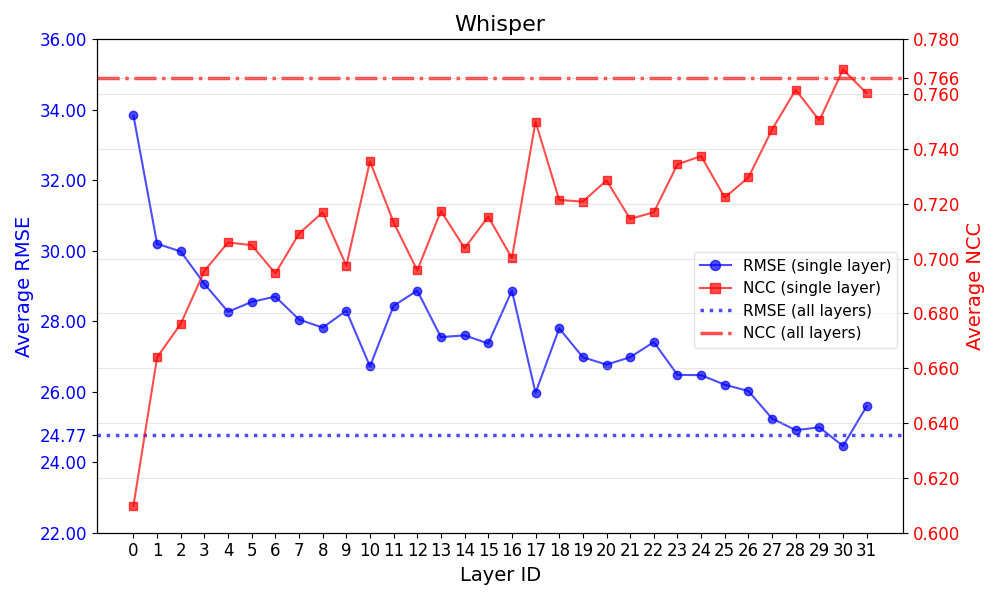}\hfill
  \includegraphics[width=.19\textwidth]{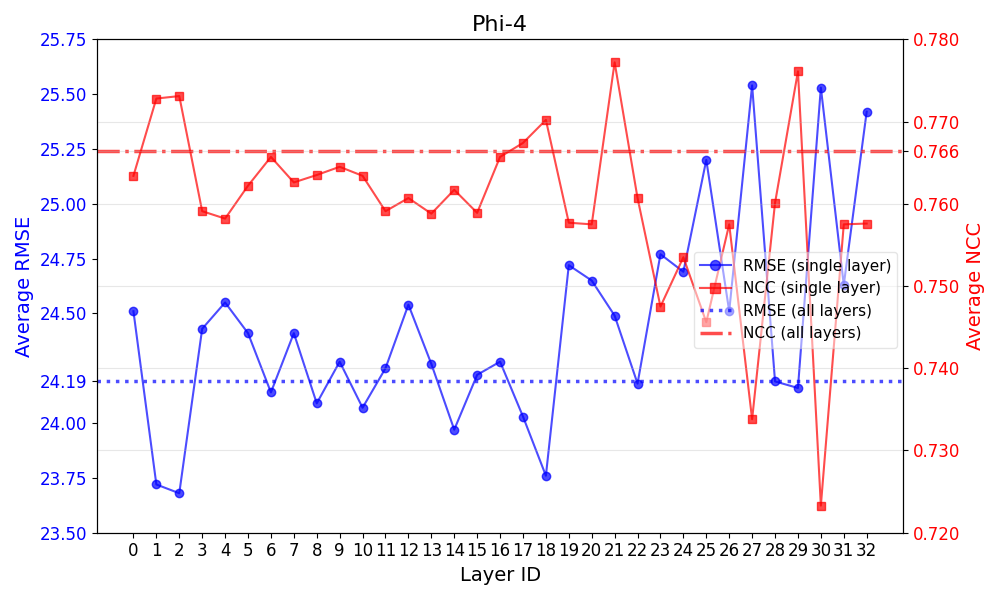} \\
  
  % Second row of 5 images
  \includegraphics[width=.19\textwidth]{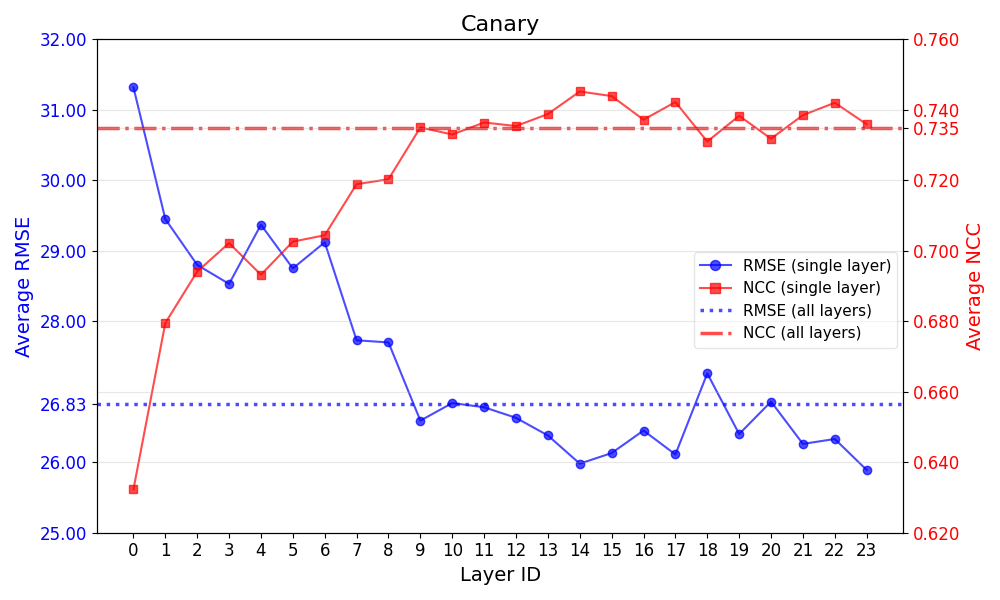}\hfill
  \includegraphics[width=.19\textwidth]{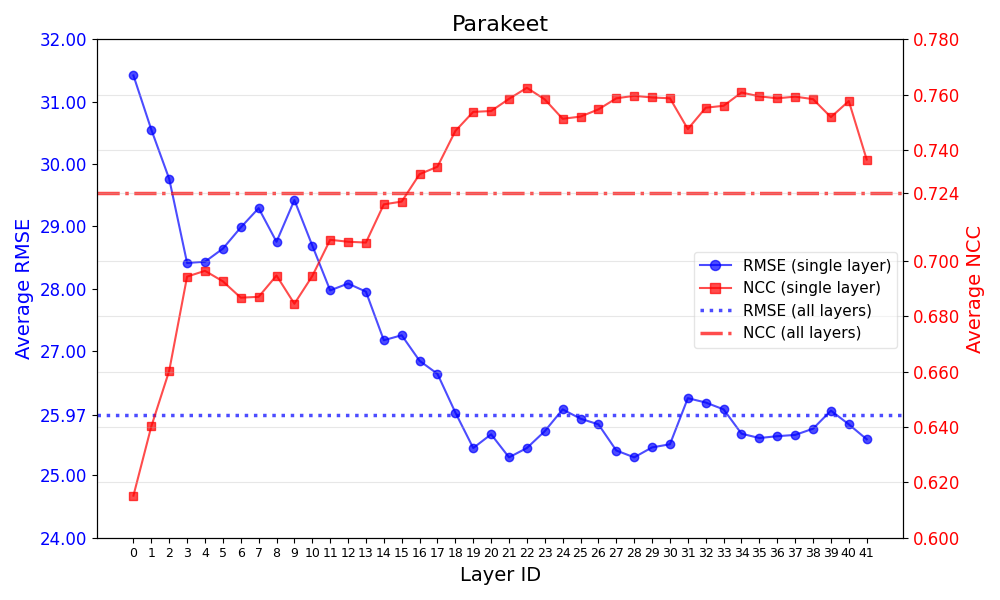}\hfill
  \includegraphics[width=.19\textwidth]{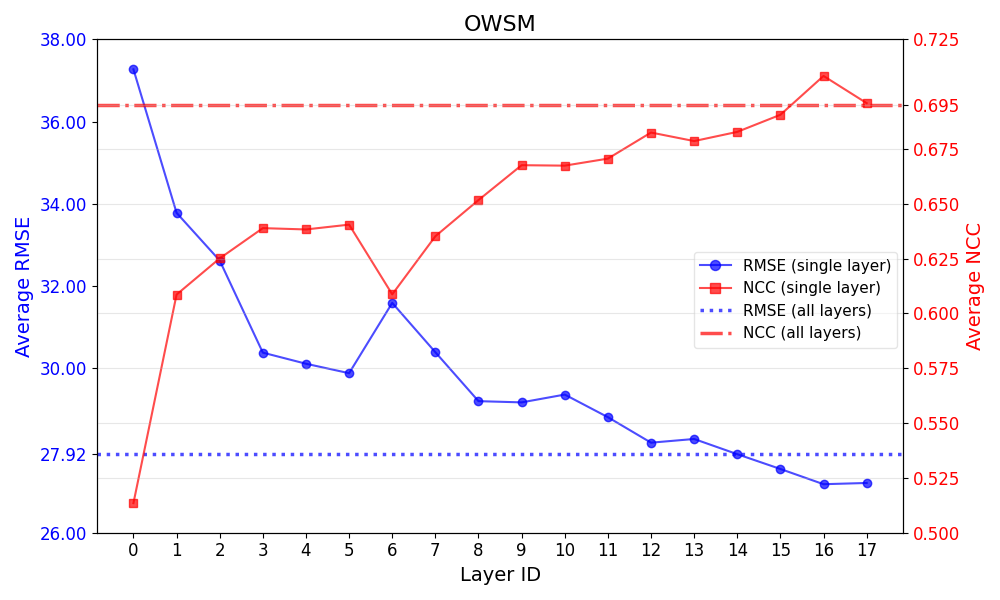}\hfill
  \includegraphics[width=.19\textwidth]{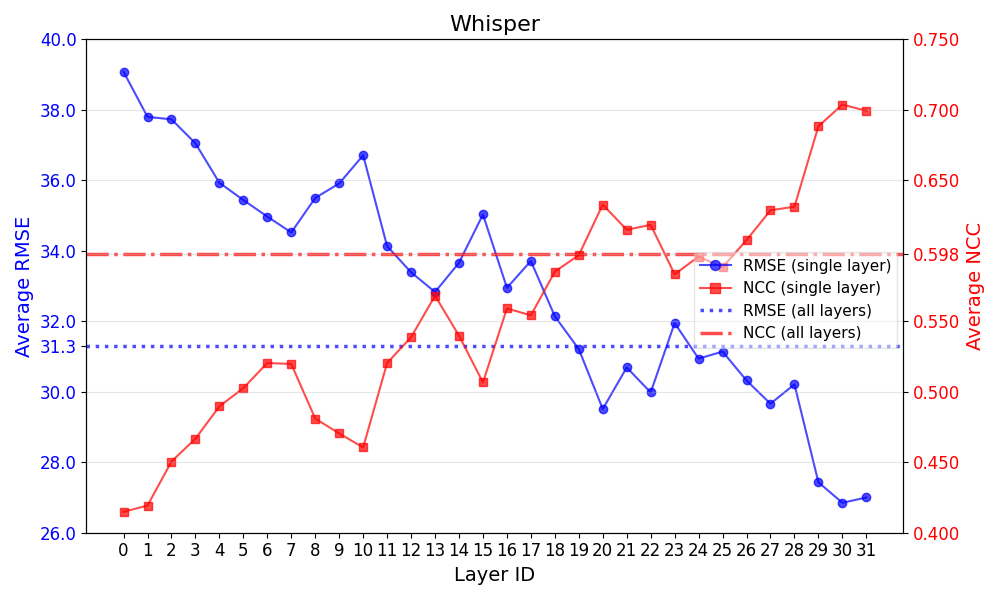}\hfill
  \includegraphics[width=.19\textwidth]{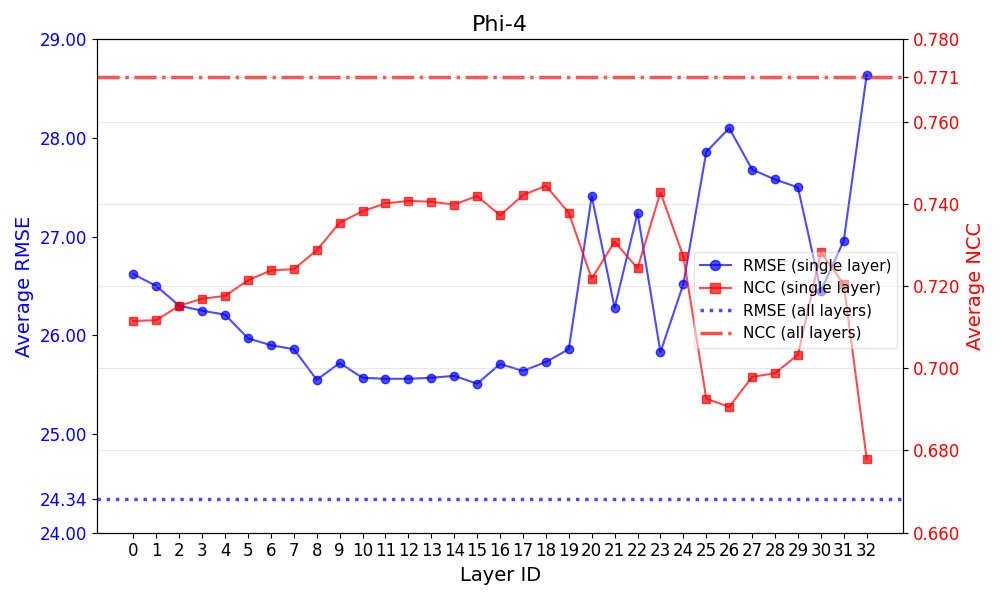}

  \caption{SIP-HI performance across different encoder layer depths or using all layers for five SFMs. The top row shows DT results, while the bottom row shows WA-TGP results. Solid lines represent results with single encoder layers while dotted lines represent results using all encoder layers. Blue and red lines indicate the average RMSE and NCC scores across the three splits.}
  \label{fig:layerwise_performance}
\end{figure*}

\subsection{Ensemble Different SFMs}
Ensembling is a widely used technique to improve model robustness and accuracy by leveraging complementary strengths of different models. To further enhance SIP-HI performance, we ensemble multiple SFMs by combining their predictions. Specifically, we take the output scores from each SFM under its best-performing configuration and learn a weighted average across models.

To explore the effectiveness of different model combinations, we evaluate all possible ensembles of 3 out of the 5 SFMs and analyze which combinations yield the strongest performance. This setup allows us to examine the interplay between model diversity and ensemble gain in the context of SIP-HI.

\renewcommand{\arraystretch}{1.15}        % optional: nicer row height

% Define a reusable “N” column type = tiny, centered, fixed‑width box
\newcolumntype{N}{>{\scriptsize\centering\arraybackslash}m{2.6cm}}

\begin{table*}[htbp]       % two‑column float
  \caption{Impact of different prediction head architectures and embedding dimensions on SIP-HI performance. Results are based on the best-performing layer configurations identified earlier. Scores for the optimal embedding dimensions for each prediction head type are in italics, while the scores for the best-performing SFM under each prediction head type are in bold.}
  \label{tab:prediction_head}
  \centering
  \setlength{\tabcolsep}{3pt}  % a bit tighter

  \begin{tabularx}{\textwidth}{p{1.05cm} *{6}{N}}
    \toprule
      \textbf{SFM}
      & \multicolumn{2}{c}{\textbf{WA‑TGP}}
      & \multicolumn{2}{c}{\textbf{WA‑TT}}
      & \multicolumn{2}{c}{\textbf{DT}}         \\
      \cmidrule(lr){2-3}\cmidrule(lr){4-5}\cmidrule(lr){6-7}
      & \textbf{RMSE} & \textbf{NCC}
      & \textbf{RMSE} & \textbf{NCC}
      & \textbf{RMSE} & \textbf{NCC}            \\
    \midrule
    Canary
      & \emph{25.85}/25.89/26.45/35.59
      & 0.730/\emph{0.736}/0.729/0.519
      & 25.25/25.32/\emph{24.56}/26.85
      & 0.764/0.727/\emph{0.772}/0.742
      & 24.65/\emph{24.33}/24.53/24.40
      & 0.765/\emph{0.771}/0.768/0.761                 \\

    Parakeet
      & 25.93/\emph{25.29}/25.72/26.01
      & 0.750/\emph{0.760}/0.751/0.751
      & 25.83/24.76/\emph{24.44}/25.97
      & 0.763/0.737/\emph{0.786}/0.761
      & 24.73/\textbf{\emph{23.56}}/25.39/26.00
      & 0.753/\textbf{\emph{0.795}}/0.758/0.760                 \\

    OWSM
      & 27.77/\emph{27.18}/27.47/36.85
      & 0.693/\emph{0.708}/0.696/0.525
      & 27.72/\emph{26.07}/26.36/28.01
      & 0.725/0.736/\emph{0.752}/0.724
      & 27.11/25.76/\emph{25.03}/27.25
      & 0.701/0.742/\emph{0.753}/0.743                 \\

    Whisper
      & 38.99/\emph{26.85}/27.17/47.01
      & 0.413/\emph{0.704}/0.689/0.202
      & 25.68/24.69/24.96/\emph{24.57}
      & 0.760/\emph{0.778}/0.767/0.774
      & 26.18/\emph{24.46}/25.26/25.73
      & 0.751/\emph{0.769}/0.765/0.753                 \\
    
    Phi-4
      & 27.35/\textbf{\emph{24.34}}/26.49/26.62
      & 0.724/\textbf{\emph{0.771}}/0.731/0.721
      & 25.56/24.80/\textbf{\emph{24.12}}/24.71
      & 0.767/0.772/\textbf{\emph{0.789}}/0.775
      & 24.91/\emph{23.68}/24.20/24.55
      & 0.739/\emph{0.773}/0.769/0.770                 \\
    \bottomrule
  \end{tabularx}
\end{table*}

\subsection{Training and Evaluation Details}
In all experiments, the SFMs are frozen, and only the prediction heads are trained. We use a batch size of 128, apply Huber loss, and train for 50 epochs. For WA-TGP, we apply a cosine annealing scheduler with an initial learning rate of 1e‑4 and a minimum learning rate of 1e-6. For transformer-based predictors, we use the same scheduler with an initial rate of 3e-5 and a linear warm-up for the first 10 epochs (start factor 0.1) and a minimum rate of 1e‑6. Optimization is performed with Adam, where $\beta_1 = 0.9$ and $\beta_2 = 0.98$.

The evaluation metrics we choose are the root mean square error (RMSE) and normalized Pearson correlation coefficient (NCC).

For each configuration, we select the checkpoint with the lowest RMSE value on the validation set for final evaluation on the test set.

\section{Results}
\subsection{Self-Learned Layer Fusion Gives Suboptimal Solutions}
\Cref{fig:layerwise_performance} presents the results of using encoder features extracted from individual layers or from all layers combined, evaluated with the WA-TGP and DT prediction heads respectively. Across both architectures and all five SFMs, except Phi-4 with WA-TGP prediction head, none of the best-performing configurations rely on features aggregated from all encoder layers. This suggests that, regardless of whether a simple learned weighting or a more complex attention-based fusion mechanism is employed, the model struggles to effectively leverage information from all layers simultaneously.

Moreover, while the best-performing layer indices for the WA-TGP and DT tend to follow similar trends within the same SFM, there is no consistent pattern across different SFMs. Specifically, Canary and Parakeet achieve their best performance using middle layers, OWSM and Whisper favor later layers, while Phi-4 performs best with earlier layers, though the performance variation across layers is relatively small in this case.

These results indicate that the optimal encoder layer for SIP-HI is model-specific and does not generalize across SFMs. Therefore, layer-wise evaluation remains essential when adapting a new SFM to this task.

\subsection{Temporal Modeling Is Important for Prediction Heads}
\Cref{tab:prediction_head} presents the results of using prediction heads with varying levels of complexity. The findings suggest that a model's ability to effectively capture temporal information is critical: both WA-TT and DT consistently outperform the simpler WA-TGP approach. However, since most SFMs achieve their best performance using features from a single encoder layer, the ability to fuse information across layers—in this case, the SFM encoder features and audiogram features—appears to play a relatively less significant role in overall SIP-HI performance.

Additionally, we note that the WA-TGP method occasionally yields extremely poor results, indicating that simpler architectures may be more sensitive to hyperparameter choices and require more careful tuning.

We also observe a significant impact of model dimensionality on performance. While there is no clear trend indicating the optimal dimension, configurations with an embedding size of 1536 tend to perform the worst. This suggests that setting the embedding dimension significantly larger than the original SFM encoder feature dimension (typically 1024) may be suboptimal.

\subsection{Stronger Individual SFMs Lead to Better Ensembles}
\Cref{tab:sfm_ranking} and \cref{tab:ensemble_ranking} presents RMSE-ranked performance of the best-performing configuration for each individual SFM used in ensembling, along with the ranked results of all possible 3-SFM ensemble combinations. The results show that all ensemble configurations outperform the best single-SFM baseline, particularly in terms of NCC, indicating that ensembling consistently enhances robustness and correlation with ground truth.

Moreover, regarding how to select ensembling candidates, the top-performing ensembles tend to include the strongest individual SFMs. For example, combinations containing Phi-4, Parakeet, and one additional SFM occupy the top three ranks in both RMSE and NCC. In contrast, the ensemble of Canary, OWSM, and Whisper, each with relatively weaker standalone performance, ranks at the bottom, reinforcing the importance of selecting strong individual models for ensembling.

To further analyze how individual SFMs contribute to the ensemble, we visualize the distribution of neural network-assigned weights used in the final ensemble predictor, as shown in \cref{fig:violin_plot}. This reveals which models the ensemble relies on most during inference. Overall, the assigned weights do not vary dramatically across SFMs, but better-performing individual models generally receive higher weights, indicating their greater influence in the final prediction. Notably, OWSM, the weakest individual performer, consistently receives lower weights, indicating its limited contribution to the ensemble output.

\begin{table}[htbp]
  \centering
  \caption{Scores of the best-performing configurations for individual SFMs, ranked by RMSE.}
  \label{tab:sfm_ranking}
  \setlength{\tabcolsep}{12pt}        % Increase column space
  \begin{tabular}{cccc}
    \hline
    \textbf{Rank} & \textbf{SFM} & \textbf{RMSE↓} & \textbf{NCC} \\
    \hline
    1 & Parakeet & 23.56 & 0.795 \\
    2 & Phi-4 & 23.68 & 0.773 \\
    3 & Canary & 24.33 & 0.771 \\
    4 & Whisper & 24.46 & 0.769 \\
    5 & OWSM & 25.03 & 0.753 \\
    \hline
  \end{tabular}
\end{table}

\begin{table}[htbp]
  \centering
  \caption{Scores of 10 three-SFM ensembles, ranked by RMSE.}
  \label{tab:ensemble_ranking}
  \setlength{\tabcolsep}{12pt}        % Increase column space
  \begin{tabular}{cccc}
    \hline
    \textbf{Rank} & \textbf{SFM Combination} & \textbf{RMSE↓} & \textbf{NCC} \\
    \hline
    1 & (Parakeet, OWSM, Phi-4) & 22.29 & 0.840 \\
    2 & (Canary, Parakeet, Phi-4) & 22.32 & 0.839 \\
    3 & (Parakeet, Whisper, Phi-4) & 22.41 & 0.839 \\
    4 & (Canary, Parakeet, OWSM) & 22.64 & 0.837 \\
    5 & (Canary, OWSM, Phi-4) & 22.65 & 0.836 \\
    6 & (Canary, Whisper, Phi-4) & 22.69 & 0.835 \\
    7 & (Canary, OWSM, Whisper) & 22.79 & 0.834 \\
    8 & (Canary, Parakeet, Whisper) & 22.80 & 0.833 \\
    9 & (OWSM, Whisper, Phi-4) & 22.97 & 0.829 \\
    10 & (Canary, OWSM, Whisper) & 23.15 & 0.828 \\
    \hline
  \end{tabular}
\end{table}

\begin{table*}[hbp]
  \centering
  \caption{{Key Attributes of the Five SFMs. The ASR WER (word error rate) is taken from the Open ASR Leaderboard\cite{open-asr-leaderboard}, focusing on English transcription. Other attributes are derived from the respective paper of each SFM. For the architecture, only the encoder and its proposed date are listed. ASR: automatic speech recognition, AST: automatic speech translation, LID: language identification, US: utterance segmentation, VAD: voice activity detection.}}
  \label{tab:attributes}
  \setlength{\tabcolsep}{8pt}  % Increase column space
  
  \begin{tabular}{ccccc}
    \hline
    \textbf{SFM} & \textbf{ASR WER} & \textbf{Data Size (hrs)}  & \textbf{Arch and Date}  & \textbf{Training Task Num} \\
    \hline
    Canary & 6.50\% & 86K & FastConformer (2023.09) & 2 (ASR, AST)  \\
    Parakeet    & 7.01\% & 64K & FastConformer (2023.09) & 1 (ASR)    \\    
    OWSM   & 7.70\% & 180K & E-Branchformer (2022.10) & 4 (ASR, AST, LID, US)      \\
    Whisper  & 7.44\% & 5M & Conformer (2020.05) & 4 (ASR, AST, LID, VAD)    \\
    Phi-4     & 6.14\% & 2M & Transformer (2017.06) & 1 (ASR)\\
    \hline
  \end{tabular}
\end{table*}

\begin{figure}[t]
  \centering
  \centerline{\includegraphics[width=\columnwidth]{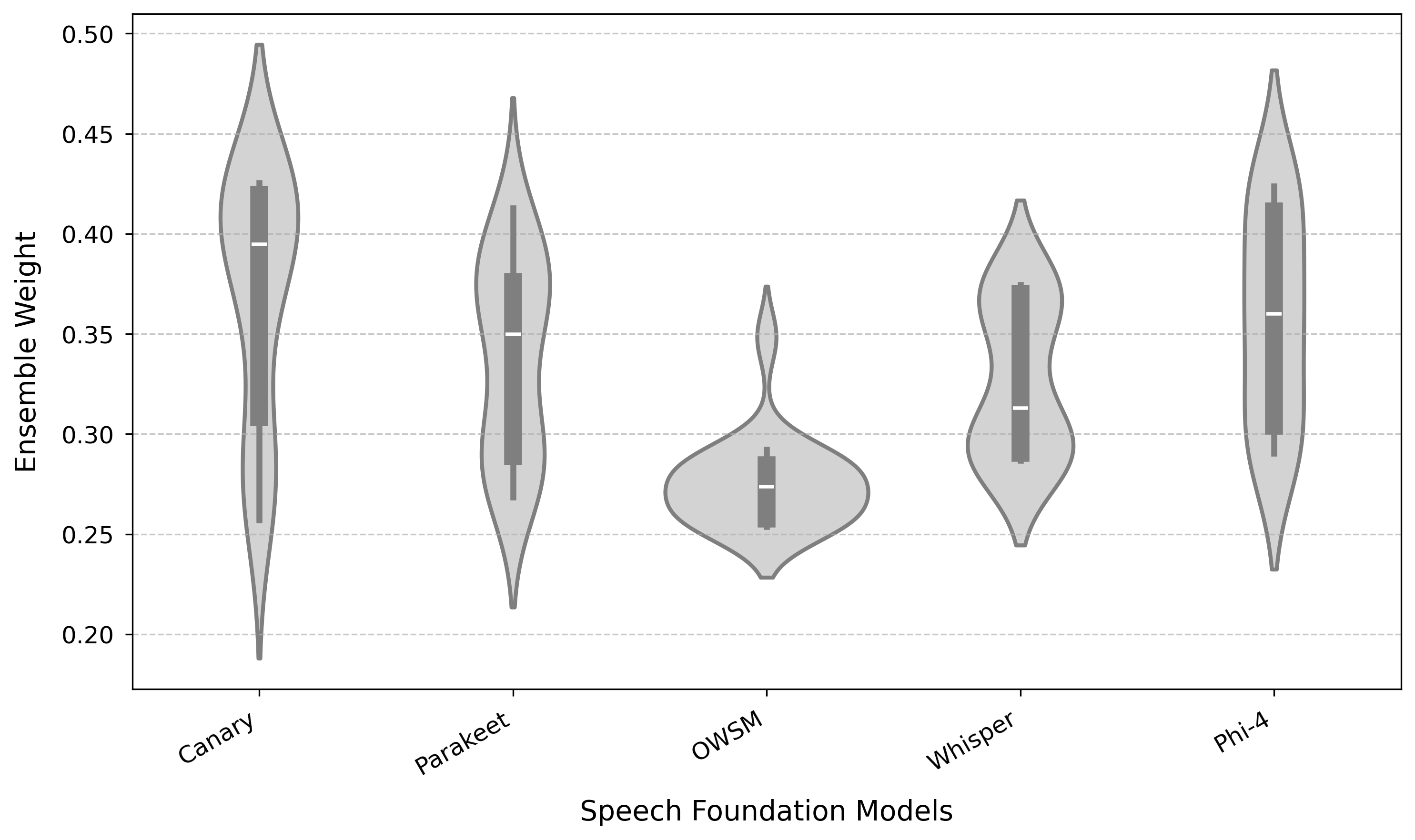}}
  \caption{Violin and box plots showing the distribution of ensemble weights assigned to each SFM. The weights are output from a softmax layer, ensuring the sum in each ensemble equals 1.}
  \label{fig:violin_plot}
\end{figure}

\begin{figure}[t]
  \centering
  \centerline{\includegraphics[width=0.8\columnwidth]{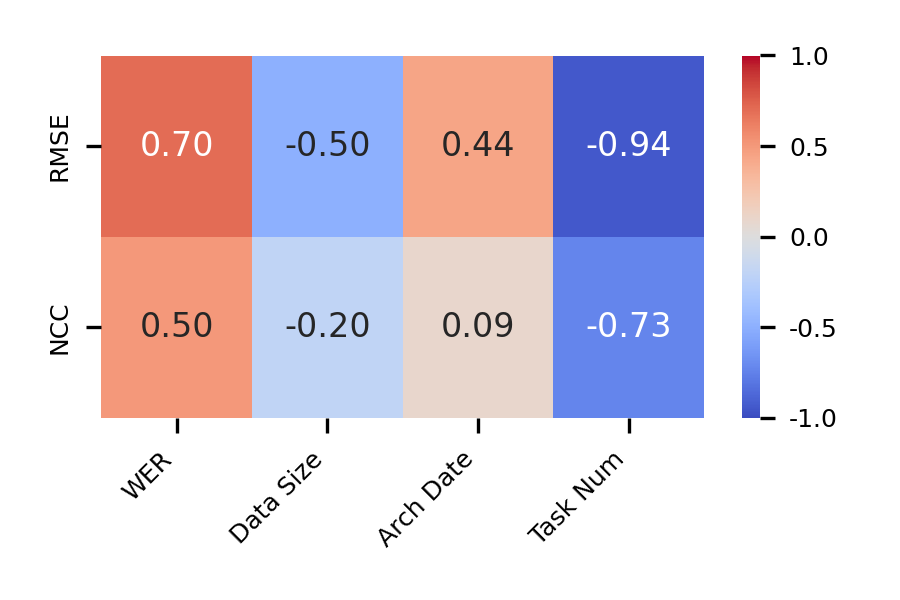}}
  \caption{Correlation between ranked SFM attributes and SIP-HI performance. Lower WER, larger data size, newer architecture, and more training tasks correspond to higher ranks when sorting.}
  \label{fig:correlation}
\end{figure}

\section{Discussion}
While it is challenging to determine which SFM performs best when adapted to SIP-HI due to the numerous variables involved in training the foundation models, we aim to explore potential relationships between SFM attributes and their SIP-HI performance. \Cref{tab:attributes} summarizes the key attributes of the five SFMs used in this study, and \cref{fig:correlation} illustrates the correlation between the ranked attributes and the SIP-HI performance rankings. We observe a strong positive correlation between individual model performance on SIP-HI and their ASR performance, as well as a strong negative relationship with the number of training tasks, suggesting that models trained exclusively on ASR tend to perform better. Additionally, models with newer architectures generally show improved performance, while surprisingly, models trained on fewer data hours achieve better results.

\section{Conclusion}
\label{sec:format}

In this paper, we present a comprehensive comparison of design choices for adapting SFMs to achieve optimal performance on SIP-HI, based on experiments on the CPC dataset. Our results highlight the significance of sweeping SFM encoder layers, designing effective temporal modeling prediction heads, and ensembling strong individual models. We hope our findings offer a framework for designing models as more SFMs emerge and provide insights for further advancements in speech intelligibility prediction for hearing-impaired individuals.

% -------------------------------------------------------------------------
% Either list references using the bibliography style file IEEEtran.bst

\clearpage
% The \IEEEtriggeratref{XX} command can be used to move to the next column before the XX-th reference
% to balance the two columns of the reference section
% \IEEEtriggeratref{XX}
\bibliographystyle{IEEEtran}
\bibliography{refs25}
% or list them by yourself:
% \begin{thebibliography}{1}

% \bibitem{waspaaweb}
% {WASPAA Website}, \url{http://www.waspaa.com}.

% \bibitem{IEEEXploreReqs}
% {IEEE {X}plore {R}equirements}, \url{https://conferences.ieeeauthorcenter.ieee.org/write-your-paper/meet-ieee-xplore-requirements/}.

% \bibitem{eWilliams1999}
% E.~Williams, \emph{Fourier Acoustics: Sound Radiation and Nearfield Acoustic Holography}.\hskip 1em plus 0.5em minus 0.4em\relax London, UK: Academic Press, 1999.

% \bibitem{cJones2003}
% C.~Jones, A.~Smith, and E.~Roberts, ``A sample paper in conference proceedings,'' in \emph{Proc. ICASSP}, vol.~II, Apr. 2003, pp. 803--806.

% \bibitem{aSmith2000}
% A.~Smith, C.~Jones, and E.~Roberts, ``A sample paper in journals,'' \emph{IEEE Trans. Signal Process.}, vol.~62, pp. 291--294, Jan. 2000.

% \end{thebibliography}

\end{document}